\documentclass{article}

\usepackage{arxiv}
\usepackage{cite}
\usepackage{amsmath,amssymb,amsfonts}
\usepackage{algorithmic}
\usepackage{graphicx}
\usepackage{textcomp}
\usepackage{xcolor}
\usepackage{enumitem}
\usepackage{tabularx}
\usepackage{booktabs}
\usepackage{pifont}
\usepackage{siunitx}
\usepackage{float}
\usepackage{hyperref}   
\usepackage{fancyhdr}
\title{LunarDepthNet: Generation of Digital Elevation Models using Deep Learning and Monocular Satellite Images}

\author{
Aaranay Aadi \\
School of Computer Science and Engineering\\
Manipal University Jaipur\\
Jaipur, India \\
\texttt{aaranayaadi@gmail.com}
\And
Jai Gopal Singla \\
Space Applications Centre (SAC)\\
Indian Space Research Organisation (ISRO)\\
Ahmedabad, India \\
\texttt{jaisingla@sac.isro.gov.in}
\And
Amitabh \\
Space Applications Centre (SAC)\\
Indian Space Research Organisation (ISRO)\\
Ahmedabad, India \\
\texttt{amitabh@sac.isro.gov.in}
\And
Nitant Dube \\
Space Applications Centre (SAC)\\
Indian Space Research Organisation (ISRO)\\
Ahmedabad, India \\
\texttt{nitant@sac.isro.gov.in}
\And
Praveen Kumar Shukla \\
Department of IoT and Intelligent Systems\\
Manipal University Jaipur\\
Jaipur, India \\
\texttt{praveen.shukla@jaipur.manipal.edu}
\And
Vijaypal Singh Dhaka \\
Department of Computer and Communication Engineering\\
Manipal University Jaipur\\
Jaipur, India \\
\texttt{vijaypalsingh.dhaka@jaipur.manipal.edu}
}

\begin{document}
\date{}

\maketitle
\begingroup
\renewcommand{\thefootnote}{}
\footnotetext{
\textit{This work has been submitted to the IEEE for possible publication. Copyright may be transferred without notice, after which this version may no longer be accessible.}
}
\endgroup

\begin{abstract}
Recent times have seen an increase in demand of high quality Digital Elevation Models (DEMs) for the lunar surface, because they are highly important for studying the moon and planning future missions. However, there is an evident lack of detailed elevation data on the Moon. To overcome this limitation, this study proposes a novel deep learning method that estimates and generates a surface elevation map directly from monocular images of the surface. The dataset used comprises of the Chandrayaan-2 Terrain Mapping Camera (TMC) images with their corresponding Digital Terrain Models (DTMs). The study proposes LunarDepthNet, which comprises of a UNet architecture to generate DEMS. It incorporates an EfficientNet encoder and custom layers to correctly learn how the light shadows on the surface relate to the actual elevation values. A combined loss function was also utilized to keep the terrain details accurate and smooth. During validation, the model showed a stable loss convergence of 12\%. It achieved a mean nRMSE of 0.437 and an MAE of 4.5m in the testing stage. These results prove the model can generate dependable elevation maps from single orbital images, which are quite useful in regions of the moon where stereo-images are not available.\\

keywords - Chandrayaan-2, Deep Learning (DL), Digital Elevation Model (DEM), Digital Terrain Model (DTM), EfficientNet, Lunar Images, UNet
\end{abstract}

\section{Introduction}
High-resolution Digital Elevation Models (DEMs) are the primary tools used for accurate measurements of the lunar surface. Recently, lunar exploration has been shifting toward precise landings and long-term stays. These models are the main tool used for navigating difficult terrain \cite{b1}. High-quality DEMs make it possible to map out surface features in detail to help with both geological studies and the practical side of mission planning. This includes everything, such as finding subtle height changes on flat plains, and measuring the extreme changes near steep crater edges\cite{b2}. With this level of data, researchers can piece together the Moon's structural past, while engineers can make sure that landers stay upright in dangerous areas.\\

Because these models are so important to mission success, reliably building high-resolution DEMs remains one of the most significant challenges in lunar remote sensing today. Even though modern missions have sent back thousands of high-quality orbital photos, turning those flat 2D images into accurate 3D maps is still a difficult technical task. All of the mission steps including landing on the moon and planning hazard-free routes, are heavily reliant on the availability of accurate features in the elevation models \cite{b3}. Even a small mistake in height measurements can lead to huge errors when calculating the steepness of a slope, which is why getting the elevation modeling right is an absolute must for any successful lunar mission.\\

Furthermore, there are difficulties in height mapping across the lunar surface because of factors like harsh lighting changes and uneven brightness \cite{b4}. These conditions are specific to the Moon and demand measurement methods that give accurate results even when visibility is poor. Better height models help make future missions safe and give researchers the clarity they need to study the geological conditions that shaped the lunar surface. By improving these techniques, it can be better understood how the Moon has changed over time and make sure that the equipment can navigate through rough lunar regions without running into trouble.

\section{Related Works}

\subsection{Recent Deep Learning frameworks}
\subsubsection{Terrestrial and Global DEMs}
Recent times have seen researchers focusing on creation of models that can turn two-dimensional images into three dimensional maps which include spatial features of a given surface. Madani \textit{et al.} \cite{b5} used generative methods to produce height maps from simple satellite photos, showing that these tools can learn the relation between surface lighting and physical terrain. For studies on Earth, Zhao \textit{et al.} \cite{b6} built a method for surface reconstruction using ZY-3 satellite data, therefore providing designs that stay reliable even when environmental conditions change. To overcome the restricted focus area of traditional convolutional designs, Yang \textit{et al.} \cite{b7} introduced a wide-range attention approach for planetary modeling. This method tracks broad spatial relationships across the landscape, which ensures that large-scale features remain consistent across the final topographic product.\\

\subsubsection{Monocular Lunar and Martian Reconstruction}
The lack of detailed ground-truth data and harsh illumination conditions make planetary topography modeling particularly difficult. As reviewed by Chen \textit{et al.} \cite{b11}, the field is moving away from purely physical models, such as Shape-from-Shading (SfS), toward hybrid or fully data-driven methods.

\begin{itemize}
    \item \textbf{Lunar Surface Reconstruction:} Majority of the recent research on Moon uses Lunar Reconnaissance Orbiter Camera (LROC) Narrow Angle Camera (NAC) images. Chen \textit{et al.} \cite{b8} created pixel-resolution DEMs by combining DL and SfS. Physical restrictions are used to modify the initial estimate provided by the network. To further enhance the resolution, Liu \textit{et al.} \cite{b9} and Chen \textit{et al.} \cite{b15} used GANs, or Generative Adversarial Networks. These models are capable of producing high-frequency details, but in order to preserve structural accuracy, they usually need a low-resolution DEM (such as an SLDEM - Digital Elevation Model co-registered with SELENE data) as a separate input during inference.
    \item \textbf{Rover and Martian Applications:} Zhong \textit{et al.} \cite{b10} enhanced local feature extraction for Yutu-2 rover PCAM pictures, which in contrast to orbital views, were captured on the surface. This enabled more precise image matching for surface reconstruction. Furthermore, Tian \textit{et al.} \cite{b13} investigated 3-D semantic terrain reconstruction from monocular close-up pictures in the Martian environment, emphasizing the usefulness of combining semantic labeling with elevation estimate.
\end{itemize}

\subsubsection{Constraint-Driven and Efficient Frameworks}

Researchers have implemented a number of constraints to guarantee that deep learning outputs continue to be geologically dependable. To ensure that pixel-scale outputs did not deviate from established low-resolution benchmarks, Chen \textit{et al.} \cite{b14} used SLDEM restrictions for large-area topography retrieval from LROC NAC pictures. In the same way, Chen \textit{et al.} \cite{b15} used the fractal characteristics of lunar landscapes to direct the generation process and developed a terrain self-similarity constraint for GANs. Efficiency is essential for practical deployment; Chen \textit{et al.} \cite{b12} created ELunarDEMNet, a lightweight architecture intended for high-throughput reconstruction of high-resolution DEMs from single-view orbiter data.\\

Modern lunar modeling often integrates multiple data sources to reconstruct surface features \cite{b9, b14, b15}. These methods use models of lower resolution, such as SLDEM2015, to correct the spatial alignment of the generated DEMs. This creates a "base-data bottleneck" where the final precision is dependent on the quality of the dataset used. Additionally, hybrid Shape-from-Shading (SfS) \cite{b8} techniques can often use repetitive optimization, which leads to excessive computational overhead on the systems on which they are run. This causes the users to rely on supplementary datasets that are either not complete in length or lack details. Because of this, the success of these elevation models ends up depending more on whether old topographic surveys already exist than on the quality of the actual photos being used.

\subsection{Contribution}

Our method treats DEM generation as a direct mapping from images to elevation. While ground-truth data supervised the training process, the model did not require elevation inputs to function once trained. By producing relative height maps from imagery alone, this approach removes the need for secondary datasets and resolves common data-access bottlenecks. Furthermore, a pathway was provided for absolute elevation recovery through a simple linear rescaling post-processing step, maintaining the accuracy of prior-dependent models while significantly reducing architectural complexity and input requirements.\

The primary contributions of this paper are: \begin{itemize} \item A lunar DEM generation framework using monocular satellite imagery. \item A UNet architecture with EfficientNet and Squeeze-and-Excitation attention tailored for lunar terrain. \item A composite loss formulation that jointly enforces pixel-wise accuracy, structural consistency, and terrain smoothness. \item A detailed evaluation demonstrating high accuracy on real lunar data. \end{itemize}

The rest of this article details the methodology, followed by the results, discussion, and conclusion.

\section{Methodology}

\begin{figure*}[h]
    \centering
    \hspace{-0.1cm}
    \includegraphics[width=0.98\textwidth]{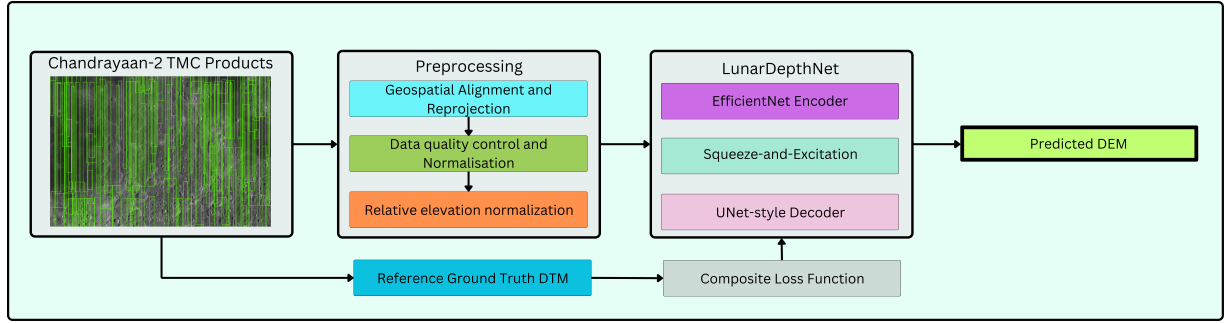}
    \caption{Overall workflow of the proposed lunar DEM generation approach. To predict surface elevation, a single-view lunar image is preprocessed and run through LunarDepthNet. While inference depends only on input imagery, reference DTMs are used solely for loss computation during training.}
    \label{fig:overflow}
\end{figure*}

The overall workflow of the proposed DEM generation approach can be seen in Fig.~\ref{fig:overflow}. A set of images and their pair DTMs are collected from the product catalogue. The dataset is preprocessed and passed through the model for training.

\subsection{Dataset}

Training and validation were performed using Chandrayaan-2 Terrain Mapping Camera (TMC-2) images and their corresponding DEMs \cite{b16}. The TMC-2 payload aboard the Chandrayaan-2 lunar orbiter, is set up to produce stereo triplets (fore, nadir, and aft views) and panchromatic photos (0.4\si{\micro\meter}
 to 0.85\si{\micro\meter}) in 5m spatial resolution from a 100 km circular orbit around the moon for creating a detailed 3-D map of the entire lunar surface. The satellite images are stored as 16-bit TIFFs, whereas DEMs are provided as unsigned 16-bit TIFFs, preserving the elevation values. Table~\ref{tab:ch2_tmc_dataset} lists the technical specifications of the selected dataset.

\subsection{Preprocessing}
The preprocessing workflow consists of geospatial alignment, data partitioning through a sliding-window approach, and a normalization strategy to convert absolute elevation into relative mapping targets.\\
\subsubsection{Geospatial Alignment and Reprojection}
To ensure strict pixel-wise correspondence, imagery is reprojected and resampled using the Rasterio library to match the Coordinate Reference System (CRS) and spatial resolution of the reference DEMs. Moving forward, bilinear resampling is used to conserve the surface textures and shading details that are needed for finding depth from a single photo. Moreover, missing data points are found using bit-depth constants and changed to zero while the data is being prepared. This gives an ability to the model to learn from information that is error free, without losing the tiny visual details it needs to estimate height accurately.

Furthermore, in order to avoid gradient corruption during training, the preprocessing handled no-data values by finding bit-depth-specific constants (such as $-32,768$ for signed 16-bit integers) and converting them to numerical zero values during array formation.\\

\subsubsection{Sliding Window Tiling and Quality Control}
The image strips are divided into $512 \times 512$ non-overlapping tiles. to make sure the model focuses on meaningful geological features rather than empty or messy data.:
\begin{enumerate}
\item \textbf{Valid Area Ratio:} Tiles are retained only if the ratio of valid pixels reaches a threshold of $\gamma = 0.05$ to mitigate the impact of sensor artifacts.
\item \textbf{Topographic Variance:} To prevent degenerate supervision from perfectly flat terrain, tiles must satisfy a peak-to-peak (PTP) elevation requirement:
\begin{equation}
Z_{max} - Z_{min} > 1.0
\end{equation}
\end{enumerate}
This filtering produced a training corpus of approximately 180,000 image-DEM pairs, partitioned into an 80:10:10 ratio for training, testing, and validation.\\

\subsubsection{Relative Elevation Normalization}

\begin{table*}[t]
\centering
\caption{Technical Specifications and Accuracy Metrics for Selected Chandrayaan-2 TMC Dataset Products}
\label{tab:ch2_tmc_dataset}
\footnotesize
\begin{tabular*}{\textwidth}{@{\extracolsep{\fill}} l c c c c c @{}}
\toprule
\textbf{Product ID (ch2\_tmc\_ndn\_)} & \textbf{Sensor} & \textbf{Resolution} & \textbf{Date of Orbital Pass} & \textbf{Lat (Deg)} & \textbf{Lon (Deg)} \\ \midrule
20240430T0222057088\_d & TMC 2 & 5m & 30/4/2024 & 31.28 & 334.14 \\
20240616T1741163186\_d & TMC 2 & 5m & 16/6/2024 & 25.67 & 67.34 \\
20250212T0459303757\_d & TMC 2 & 5m & 12/2/2025 & -0.26 & 321.59 \\
20240715T1335557435\_d & TMC 2 & 5m & 15/7/2024 & -4.92 & 47.66 \\
20241204T2145097873\_d & TMC 2 & 5m & 4/12/2024 & -35.22 & 153.90 \\
20220602T0411113533\_d & TMC 2 & 5m & 2/6/2022 & 61.30 & 167.57 \\
20250209T0217318134\_d & TMC 2 & 5m & 9/2/2025 & -0.37 & 2.18 \\ \bottomrule
\end{tabular*}
\end{table*}

Since the main challenge countered in this study was relative mapping of values instead of capturing absolute details, elevation values were normalized on a relative scale. This was accomplished by employing the following normalizing technique:

\begin{itemize}
    \item \textbf{Image Stretching:} 
To handle the varying sun angles and the difference in sensor values across different orbits, images went through percentile stretching. Pixel intensities were clipped at the 1st and 99th percentiles, then rescaled to a range between 0 and 1. During dataset loading, these values were normalized again using a mean and standard deviation of 0.5, which centered the input data around zero for the model.
    \item \textbf{Local DEM Rescaling:} 
The target DEM tiles were locally normalized to map the physical elevation values into a unit range $[0, 1]$. For a given elevation pixel $Z_{i,j}$ within a tile, the normalized target $Z'_{i,j}$ is calculated as
\begin{equation}
    Z'_{i,j} = \frac{Z_{i,j} - Z_{min}}{Z_{max} - Z_{min} + \epsilon}
\end{equation}
where $Z_{min}$ and $Z_{max}$ are the local extrema of the tile, and $\epsilon$ is a small constant ($1e-3$) to prevent division by zero in the near-flat regions. The metadata (local $Z_{min}$ and $Z_{ptp}$) were stored during inference, allowing the relative prediction to be linearly rescaled back to absolute physical units during post-processing.\\
\end{itemize}

\iffalse
\subsection{Problem Formulation}
Given a monocular high-resolution orbital image $I \in \mathbb{R}^{H \times W}$, the objective is to predict the corresponding digital elevation model $\hat{D} \in \mathbb{R}^{H \times W}$. Unlike stereo or multimodal approaches, the proposed method relies exclusively on single-view imagery during inference. Ground-truth DEMs are used only during training for supervision and are not provided as inputs to the network.
\fi

\subsection{Network Architecture}

\begin{figure*}[h]
    \centering
    \hspace{-0.1cm}
    \includegraphics[width=0.98\textwidth]{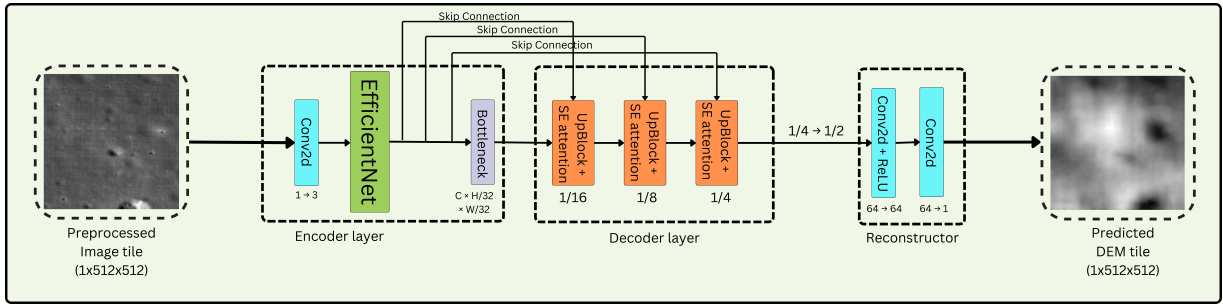}
    \caption{Layer-wise architecture of LunarDepthNet: The model uses SE-enhanced decoder blocks and an EfficientNet encoder in a UNet-style encoder-decoder architecture. Accurate lunar DEM prediction is made possible by multi-scale skip connections, which maintain spatial detail.}
    \label{fig:architecture}
\end{figure*}

\textbf{LunarDepthNet} follows a UNet-style encoder–decoder architecture with explicit multiscale feature fusion and channel-wise recalibration. The design balances representational capacity with computational efficiency, making it well suited for high-resolution lunar imagery. Fig.~\ref{fig:architecture} illustrates the architecture of the proposed model

\begin{itemize}
    \item \textbf{Encoder:} Based on EfficientNet-B3, pretrained and adapted for a single-channel grayscale input via a learnable convolutional projection to three channels. By extracting hierarchical feature maps at various spatial resolutions, the encoder captures both more general lighting context and fine-scale texture information.
    
    \item \textbf{Bottleneck:} The deepest encoder feature map is used to represent the bottleneck. Adaptive global average pooling is used to capture a compact global terrain context that preserves uniformity between local and global elevation structures.
    
    \item \textbf{Decoder:}  consists of four phases of progressive upsampling:
    \begin{itemize}
        \item transposed convolution to upsample spatially;
        \item Skip connections from corresponding encoder stages to preserve spatial detail;
        \item Non-linear activation and group normalization for stable optimization;
        \item Dropout for regularization;
        \item Squeeze-and-Excitation (SE) layers for channel-wise feature response recalibration.
    \end{itemize}
    
    \item \textbf{Output Heads:} The network uses a sigmoid-activated elevation head to generate a single-channel normalized DEM. To maintain consistency, an auxiliary scale head regresses the global elevation parameters during training based on the bottleneck characteristics. This head mostly serves as a regularizer and does not add any new inference dependencies.
\end{itemize}

\subsection{Training Parameters}

The Adam optimizer was used to train the model in a single-stage supervised fashion, with a weight decay of $1 \times 10^{-5}$ and a learning rate of $5 \times 10^{-5}$. To increase convergence stability, a cosine annealing learning rate scheduler was used. The training was done on an NVIDIA H100 GPU for 200 epochs in batches of 32. It took about six hours to complete the training. Held-out DEM tiles were used for validation in order to keep a check on generality and avoid overfitting.\\
To ensure both numerical accuracy and structural fidelity, The training used a composite loss function:
\begin{equation}
    \mathcal{L} = \alpha_{L1}\mathcal{L}_{L1} + \alpha_{\text{grad}}\mathcal{L}_{\text{grad}} + \mathcal{L}_{\text{SSIM}}
\end{equation}
where:
\begin{itemize}
    \item $\mathcal{L}_{L1}$ enforces absolute elevation accuracy,
    \item $\mathcal{L}_{\text{grad}}$ preserves terrain edges and slope continuity,
    \item $\mathcal{L}_{\text{SSIM}}$ maintains structural similarity between the predicted and reference DEMs.
\end{itemize}
All the weight losses were set to unity during the experiment.

\subsection{Inference Modes}
The proposed framework allows for two inference configurations:

\begin{itemize}
    \item \textbf{Image-Only Inference (Relative DEM):} When only orbital imagery is available, the trained model produces a relative DEM that preserves fine-scale topographic structure and elevation differences within each tile. This mode requires no auxiliary data and is suitable for large-scale surface reconstruction, navigation analysis, and comparative geomorphological studies.
    
    \item \textbf{Absolute DEM Recovery via post-hoc scaling:} If reference elevation statistics (e.g., minimum and maximum elevation values of a tile) are available, the relative prediction can be linearly rescaled to recover absolute elevation values:
    \begin{equation}
        \hat{D} = \hat{D}_{\text{norm}} \cdot \left(\max(D) - \min(D)\right) + \min(D)
    \end{equation}
    This step is applied after inference, and does not introduce additional inputs or dependencies into the network. As a result, the model remains a single-image DEM predictor while enabling optional absolute elevation recovery when reference information is available.
\end{itemize}

\section{Results}

LunarDepthNet performs exceptionally well, when it comes to DEM prediction from high-resolution lunar images. The model was tested on a variety of lunar landscapes, including areas with complicated topography including slopes, ridges, and craters. The model shows exact pixel-level matching in numerous locations, with an observed deviation of about 10~m in complicated areas, based on a quantitative evaluation against ground-truth DEMs.

\subsection{Evaluation Metrics}
To assess model performance quantitatively, the following metrics are employed:
\begin{itemize}
    \item \textbf{Mean Absolute Error (MAE):} Measures the average absolute difference between ground-truth and predicted elevation (in meters).
    
    \item \textbf{Normalized Root Mean Squared Error (nRMSE):} Represents global reconstruction accuracy while penalizing larger errors. The metric is unitless and is computed by normalizing RMSE over the elevation range. The reported mean nRMSE is obtained by averaging the per-tile nRMSE values across all processed tiles.
\end{itemize}

\subsection{Performance Evaluation} In a single-stage supervised condition, the model showed strong convergence. The final training and validation losses were roughly 9\% and 12\%, respectively, suggesting strong generalization over unseen lunar territory. The final design came from a series of small improvements to the network structure. At first, a basic UNet produced terrain that looked blurry and lacked sharp details. Switching the encoder to a ResNet50 helped define surface features better, though the average error was still higher than desired. The best performance came from using an EfficientNet-B3 backbone with squeeze-and-excitation layers and a combined loss function, which improved both small-scale accuracy and training reliability. This final version gave an \textbf{MAE of 4.5 m} and a \textbf{mean nRMSE of 0.437}. While most errors stayed within a 10-meter range, the low nRMSE confirms the model is particularly good at avoiding large vertical mistakes across various lunar landscapes. Table \ref{tab:table_results} compares the performance of iterative experiments

\begin{table}[H]
\caption{Performance Comparison of Model Iterations}
\label{tab:table_results}
\begin{center}
\small
\begin{tabular}{|l|l|c|c|c|}
\hline
\textbf{Architecture} & \textbf{Backbone} & \textbf{Val Loss} & \textbf{MAE(m)} & \textbf{nRMSE} \\
\hline
Basic UNet & Standard & 28\% & $>$ 10.0 & -- \\
\hline
ResNet-UNet & ResNet50 & 19\% & 7.2 & 0.815 \\
\hline
\textbf{LunarDepthNet} & \textbf{EffNet-B3} & \textbf{12\%} & \textbf{4.5} & \textbf{0.437} \\
\hline
\end{tabular}
\end{center}
\end{table}

\begin{figure}[htbp]
    \centering
    % Use \linewidth to ensure it scales to exactly 100% of the column width
    \includegraphics[width=1\linewidth]{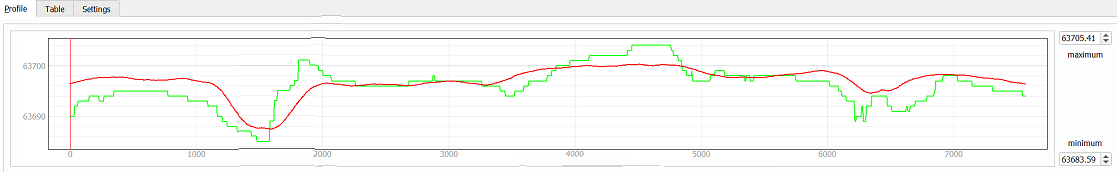}
    \caption{Comparison of relative elevation profiles showing the alignment of the predicted DEM (red) with the ground-truth DEM (green) when inference is conducted without using the original DEM.}
    \label{fig:relative_profile}
\end{figure}

\begin{figure}[htbp]
    \centering
    % Use \linewidth to ensure it scales to exactly 100% of the column width
    \includegraphics[width=1\linewidth]{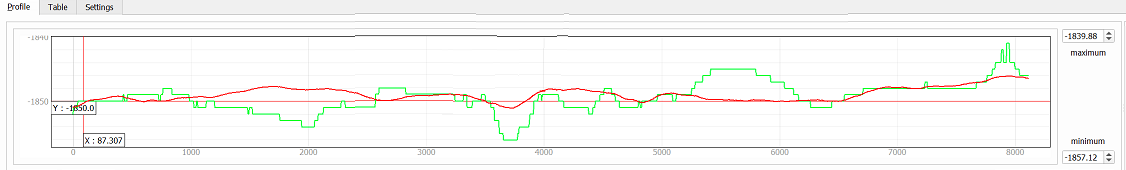}
    \caption{Comparison of absolute elevation profiles, demonstrating the post-hoc scaling technique that uses the original DEM for reference during inference to recover absolute elevation values.}
    \label{fig:absolute_profile}
\end{figure}

Fig.~\ref{fig:relative_profile} compares the relative elevation profile between a generated DEM and the original ground-truth DEM when the inference does not use the original DEM for reference. Meanwhile, Fig.~\ref{fig:absolute_profile} shows the absolute elevation profile obtained when the inference uses the original DEM for reference.

\section{Discussion}
Table \ref{tab:comparative_analysis} demonstrates how well the proposed model compares to existing elevation techniques. Most existing techniques remain limited by their heavy processing demands. In contrast, the method proposed here provides a simplified framework that maintains rigorous vertical accuracy, making it highly effective for large-scale mapping. 

Instead of the usual blurring caused by broad normalization, this model uses specific parameters to keep local features sharp. Furthermore, the network retains minor characteristics that are typically washed out, such as ridges and modest slopes, by concentrating on local height ranges.

\begin{table*}[t]
\centering
\caption{Comparative Analysis of Elevation Estimation Methods}
\label{tab:comparative_analysis}
\footnotesize
\begin{tabularx}{\textwidth}{@{}l l l l X X@{}}
\toprule
\textbf{Method/Ref} & \textbf{Domain} & \textbf{Modality} & \textbf{Type} & \textbf{Performance/Contribution} & \textbf{Key Limitations (in training/inference)} \\ \midrule
Madani et al. [5] & Earth & RGB Image & Generative & Improved visual quality from single RGB, mean nRMSE of 0.4643 & Dependence on synthetic training data \\
Zhao et al. [6] & Earth & Stereo Image & Supervised & Meter-level vertical accuracy using ZY-3, RMSE of 3.297m & Inherent stereo dependency \\
Yang et al. [7] & Planet & Imagery/Base & Attention & Global spatial consistency via attention, RMSE of 2.498m & High computational cost \\
Chen et al. [8] & Lunar & Image+Illum. & Hybrid & Pixel-resolution local accuracy, generation of 1.6m DTM & Complex physics-based SfS priors \\
Liu et al. [9] & Lunar & Image+DEM & Generative & High-frequency surface detail generation, RMSE of 3-10m & Low-res DEM dependency at inference \\
Zhong et al. [10] & Lunar & PCAM Image & Feature-DL & Enhanced rover-scale image matching & Limited to in-situ reconstruction \\
Chen et al. [12] & Lunar & Single Image & Supervised & Efficient high-resolution reconstruction, RMSE of 2.86m & Sensitivity to varying lighting noise \\
Tian et al. [13] & Mars & Close-up Image & Semantic & Semantic-aware 3D reconstruction, mean nRMSE of 0.510 & Scene-specific close-up limitation \\
Chen et al. [14] & Lunar & Image+SLDEM & Supervised & Robust large-area topography retrieval, RMSE of 1.254m & Dependence on SLDEM constraints \\
Chen et al. [15] & Lunar & Image+DEM & Generative & Improved geological consistency, RMSE of 2.01m & Requires low-res DEM base at inference \\ \midrule
\textbf{Proposed} & \textbf{Lunar} & \textbf{Single Image} & \textbf{Supervised} & \textbf{Single view imagery, MAE of 4.5m, mean nRMSE of 0.437} & \textbf{Requires base DEM for geodetic scaling} \\ \bottomrule
\end{tabularx}
\end{table*}

\section{Conclusion}
This paper presents a supervised mapping approach for generating lunar DEMs from monocular images. By incorporating Squeeze-and-Excitation attention into a UNet architecture, the model recovers detailed topographic features without requiring auxiliary data at runtime. With an MAE of 4.5 m, and a mean nRMSE of 0.437, the model performs consistently across varying lunar terrains while remaining computationally efficient. Although extreme lighting still poses a challenge, this framework offers a scalable method for high-resolution mapping and autonomous navigation. A key limitation is vertical errors of up to 10 m in extreme illumination conditions, particularly in shadowed or highly reflective regions. Future work will explore stronger backbones, such as Vision Transformers or deeper feature extractors, to improve robustness in high-contrast lunar terrains.

\iffalse
\paragraph{Inference Requirement Comparison}
Table ~\ref{tab:inference_reqs} compares the inference requirements for different elevation estimation methods. Specifically, it highlights whether the methods require original DEMs, stereo images, or physics-based constraints during inference.\\
\fi

\section*{Acknowledgement}
We would like to thank the Director of the Indian Space Research Organization's (ISRO) Space Applications Center (SAC) for giving the institutional support and permission needed to publish this study. We also acknowledge the use of the Chandrayaan-2 TMC-2 data made available through the ISSDC portal.

%\section*{References}

\end{document}